\documentclass{article}

\PassOptionsToPackage{numbers, compress}{natbib}
 \usepackage[preprint]{neurips_2026}

\usepackage[utf8]{inputenc} 
\usepackage[T1]{fontenc}    
\usepackage{hyperref}       
\usepackage{url}            
\usepackage{booktabs}       
\usepackage{amsfonts}       
\usepackage{nicefrac}       
\usepackage{microtype}      
\usepackage{xcolor}         
\usepackage{xspace} 
\usepackage{multirow}
\usepackage{amsmath}
\usepackage{algorithm}
\usepackage{algpseudocode}
\usepackage{booktabs}
\usepackage{makecell}
\usepackage{graphicx}
\newcommand{\ourmethod}{AQuaUI\xspace}

\newcommand{\showui}{\textsc{ShowUI}\xspace}

\usepackage{enumitem}

\usepackage{cleveref}
\Crefformat{figure}{#2Fig.~#1#3}
\Crefmultiformat{figure}{Figs.~#2#1#3}{ and~#2#1#3}{, #2#1#3}{ and~#2#1#3}
\Crefformat{table}{#2Tab.~#1#3}
\Crefmultiformat{table}{Tabs.~#2#1#3}{ and~#2#1#3}{, #2#1#3}{ and~#2#1#3}
\Crefformat{appendix}{Appx.~\S#2#1#3}
\Crefformat{section}{\S#2#1#3}
\Crefformat{subsection}{\S#2#1#3}
\crefformat{algorithm}{Alg.~#2#1#3}
\crefformat{equation}{Eq.~#2#1#3}
\crefformat{enumi}{Prop.~#2#1#3}
\crefformat{listing}{Code.~#2#1#3}
\crefformat{lstlisting}{Code.~#2#1#3}
\Crefmultiformat{enumi}{Prop.~#2#1#3}{ and~#2#1#3}{, #2#1#3}{ and~#2#1#3}

\title{\ourmethod: Visual Token Reduction for GUI Agents\\ with Adaptive Quadtrees}

\author{%
  Yuankai Li \\
  UC Davis\\
  \texttt{ykali@ucdavis.edu}\\
  \And
  Tinghui Zhu \\ 
  UC Davis \\
  \texttt{thuzhu@ucdavis.edu}\\
  \And
  Ha Min Son \\
  UC Davis\\
  \texttt{hmson@ucdavis.edu}\\
  \And
  Zhe Zhao \\
  UC Davis \\
  \texttt{zao@ucdavis.edu}\\
  \And
  Xin Liu \\
  UC Davis \\
  \texttt{xinliu@ucdavis.edu}\\
  \And
  Muhao Chen \\
  UC Davis \\
  \texttt{muhchen@ucdavis.edu}
}

\begin{document}

\maketitle

\begin{abstract}
Large Multimodal Models (LMMs) have recently emerged as promising backbones for Graphical User Interfaces (GUIs) agent models, where high-resolution GUI screenshots are introduced to the prompts at each iteration step. However, these screenshots exhibit highly non-uniform spatial information density: large regions may carry little information and are visually homogeneous, while key text and icons may require high visual fidelity. Existing approaches to this problem either require additional training or rely on attention-based token compression, ignoring the structured layout and spatial redundancy of GUI screenshots. To fill the gap, this paper proposes \textsc{\textbf{\ourmethod}}, a training-free inference-time token reduction method for GUI agent models that utilizes the non-uniform information density in screenshots. \ourmethod constructs an adaptive \textit{quadtree} on each screenshot input and keeps one representative merged token per leaf of the quadtree. \ourmethod preserves the spatial positions of retained tokens throughout the pipeline to ensure that all position-encoding stages remain consistent. 
To further improve temporal consistency across multi-step GUI interactions, we propose a \textit{conditional quadtree} algorithm that leverages the continuity between consecutive screenshots within a single request. Specifically, it refines the current quadtree using previous quadtrees as references, helping preserve fine-grained regions across static or mildly shifted GUI states.
We implement \ourmethod on state-of-the-art GUI agent models and conduct experiments on standard grounding and navigational benchmarks. \ourmethod consistently shows improved accuracy--efficiency trade-offs over prior baselines. Notably, on GUI-Owl-1.5-32B-Instruct, \ourmethod achieves up to \textbf{13.22\%} speedup and \textbf{29.52\%} fewer visual tokens while retaining \textbf{99.06\%} of full-token performance, suggesting that the spatial redundancy of GUI screenshots can be exploited at inference time without retraining.

\end{abstract}

\section{Introduction}

With the rapid development of Large Multimodal Models (LMMs), current state-of-the-art models have gained the ability to understand and navigate through complex UI elements in Graphical User Interfaces (GUIs). This, combined with research progress within autonomous agents, has led to various GUI agents designed to interpret user instructions and autonomously perform complex tasks. As a result, GUI agents have emerged as a new research focus of 
agentic AI.

Earlier GUI agents often adopt a hybrid way that combines textual representations, such as accessibility trees or HTML, with visual information~\citep{deng2023mind2web,zhou2023webarena,wang2024mobile,lu2024omniparser}. 
Such systems, however, often face cross-platform adaptability and require frequent task-specific optimization. 
In response to these limitations,
native GUI agent models that directly take screenshots as input~\citep{xu2024aguvis,anthropic2024computer,lin2025showui} have become increasingly popular as they 
demonstrate competent performance on trending GUI agent benchmarks ~\citep{xie2024osworld, rawles2024androidworld}. Screenshots provide a natural perspective of understanding GUIs and a unified solution in cases where accessibility trees are unavailable. 
The end-to-end design also offers a data-centric view that allows 
model performance to improve through data scaling and iterative feedback.

Such GUI agent models typically needs to process high-resolution screenshots at every interaction step, resulting in a massive number of visual tokens even with merging and patching techniques. For example, a typical Pixel 6 phone screenshot with a resolution of 2400$\times$1080, when processed with Qwen3-VL~\citep{bai2025qwen3}, can produce around 3000 tokens. 
This 
substantially increases computational cost and limits the amount of conversation history that can be retained within the context window.
Token compression, therefore, has become increasingly important for GUI agents, as it can reduce inference cost, improve throughput, and enable longer interaction histories.

Recent research 
observes that GUI screenshots contain substantially more low-information regions than natural images, while also exhibiting simpler and more regular spatial structures~\citep{huang2025gui, lin2025showui}. Previous works of token compression in Vision Language Models (VLMs) target natural images and are therefore not well suited to GUI images~\citep{chen2024image,jiang2025kind,zhang2025enhancing}. Other works that focus on GUI images develop token reduction methods as part of GUI agent model training~\citep{lin2025showui,ouyang2026focusui}, leaving open whether they can be directly applied to arbitrary GUI agent models without training.

\begin{figure*}[t]
  \centering
  \scalebox{1}[0.9]{%
        \includegraphics[width=0.98\linewidth]{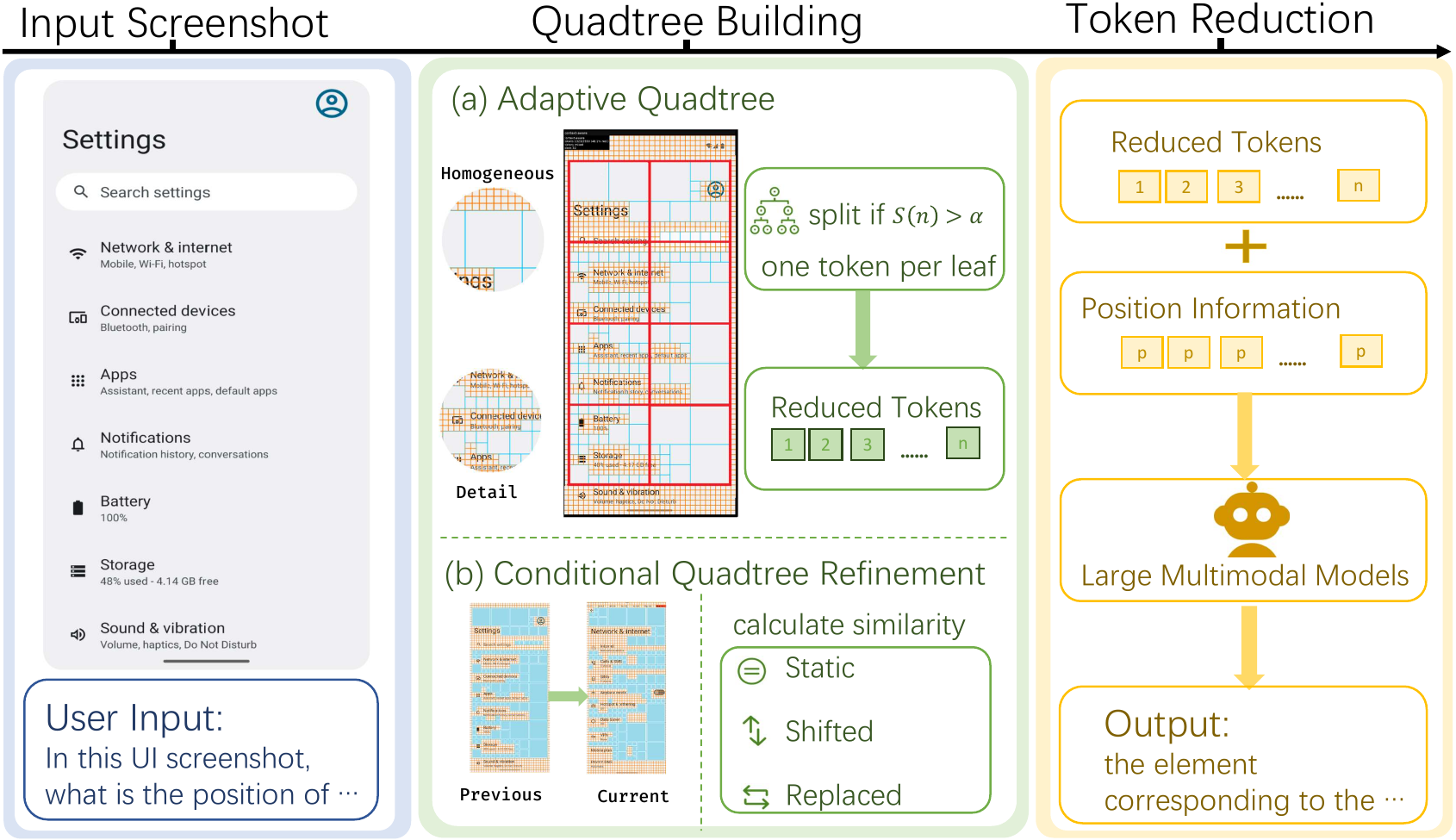}
    }
  
  \caption{An overview of the entire pipeline of \ourmethod. Given a GUI screenshot, \ourmethod will first build an adaptive quadtree to capture the UI layout. For each leaf node in that quadtree, all but one representative tokens are discarded. These representative tokens are sent to the languge model with their position information for the final output. Additionally, we design a conditional quadtree refinement algorithm that reuses previous screenshots when building the quadtree to remain token consistency.}
  \label{fig:teaser}
  \vspace{-2mm}
\end{figure*}

To fill this gap, we propose \ourmethod, a training-free framework that can be easily applied to any downstream LMMs. We use \textit{quadtree}~\citep{4766900} to select certain tokens before they are send to the text transformer. 
The pipeline is illusatrated in \Cref{fig:teaser}.

We evaluate \ourmethod on grounding benchmarks including UI-Vision~\citep{nayak2025ui}, ScreenSpot-Pro, ScreenSpot-V2~\citep{li2025screenspot}, OSWorld-G~\citep{xie2024osworld} and MMBench-GUI~\citep{wang2025mmbench}. For navigational evalution, we adopt AndroidWorld~\citep{rawles2024androidworld} and AndroidControl~\citep{li2024effects}. \ourmethod can outperform prior baselines with better accuracy-latency trade-off, especially with larger models. On Qwen3-VL models, \ourmethod can compress the visual tokens by 30\% without hurting overall performance.

In summary, the main contribution of our paper can be summarized as follows:
\begin{itemize}[leftmargin=1.5em]
    \item We propose \ourmethod, a training-free inference-time visual-token reduction method that exploits the structural layout of GUI screenshots using an adaptive quadtree.
    \item We introduce a conditional quadtree algorithm that uses previous screenshots within the same request to preserve fine-grained partitions across static or mildly shifted screenshots.
    \item We implement \ourmethod on various state-of-the-art GUI agent models and evaluate it on standard GUI grounding and navigation benchmarks, showing improved accuracy--efficiency trade-offs over prior baselines.
\end{itemize}

\section{Related Work}
\subsection{GUI Agents}






Recent work on GUI agents has rapidly expanded both the scale of
backbone models and the range of supported platforms. UI-TARS~\citep{qin2025ui} and GUI-Owl~\citep{ye2025mobile} train LMMs on screen-grounded trajectories, while UI-Voyager~\citep{lin2026ui}, MAI UI~\citep{zhou2025mai}, and ClawGUI~\citep{tang2026clawgui} explore complementary directions in planning, multi-agent coordination, and tool use. Despite various differences, these systems share a common serving bottleneck: each interaction step introduces high-resolution screenshots whose visual tokens can dominate prefill latency, attention cost, and KV-cache memory.

Several recent methods recognize that UI screenshots require specialized treatment. ShowUI~\citep{lin2025showui} reduces visual tokens by leveraging UI structural regularity, while FocusUI~\citep{ouyang2026focusui} selects instruction-relevant UI patches and introduces a position-aware strategy to mitigate the loss of positional continuity. These methods highlight the importance of spatial structure for UI grounding, but their main results are obtained in model-specific training settings. In contrast, AQuaUI targets training-free deployment on off-the-shelf models.

\subsection{Compression Methods for LMMs}
\textbf{Token Pruning.} A broad line of work reduces the number of visual tokens by estimating token importance from attention and similarity. FastV~\citep{chen2024image} prunes visual tokens in deep layers using attention from the language stream; G-Prune~\citep{jiang2025kind}
models token importance through graph propagation; and SimIgnore~\citep{zhang2025enhancing} uses cross-modal similarity to
discard redundant patches. Recent generic pruners such as HiPrune~\citep{liu2026hiprune} and VisionZip~\citep{yang2025visionzip} further improve training-free visual-token reduction.
These methods primarily frame reduction as token-importance estimation. GUI screenshots introduce a different situation where low-attention tokens may still be important because they define labels or coordinate context. Equally importantly, such methods rely on explicit calculation of attention maps, which could add extra complexity to FlashAttention~\citep{dao2022flashattention} or KV Cache management.





\textbf{KV-cache compression.} 
Another line of work compresses the KV cache after tokens have already entered the transformer. SnapKV~\citep{li2024snapkv} selects KV entries
using attention from an observation window, while
PyramidKV~\citep{cai2024pyramidkv} allocates different cache budgets across different layers. Subsequent
methods improve selection with diversity terms, pre-RoPE scores, or mixed policies~\citep{cai2025r,mao2026triattention}. GUI KV~\citep{huang2025gui} specifically calculates the spatial and temporal scoring of KV cache based on GUI screenshots. These techniques are largely orthogonal to \ourmethod:
they reduce memory and decoding cost after visual tokens have been encoded, whereas \ourmethod reduces the visual token sequence itself before KV-cache is computed.




\section{Method}
In this section, we formalize the problem in \Cref{method-def}, then present the basic token reduction method for \ourmethod in \Cref{method-quad}. Building a conditional quadtree using previous images is discussed in \Cref{method-cond}. 

\subsection{Task Definition}
\label{method-def}

At each step $t$, a GUI agent receives a screenshot $s_t$ and a text prompt $p_t$ containing the user instruction and interaction history. A native GUI agent model produces a response and action $(r_t,a_t)=\mathcal{M}(p_t,\{s_i\}_{i=0}^t)$. 
Each screenshot $s_t$ is encoded as a sequence of visual tokens $\text{tok}_{\text{vis}}$ using a vision encoder, which are processed jointly with text tokens $\text{tok}_
{\text{text}}$. The goal is to find a different set of tokens $\text{tok}_{\text{comp}}$ whose length $|\text{tok}_{\text{comp}}| < |\text{tok}_{\text{vis}}|$ while preserving task performance.


\subsection{Adaptive Quadtree Tokenization}
\label{method-quad}
For GUI screenshots, information is particularly  non-uniformly distributed~\citep{lin2025showui}, and we propose a \textit{quadtree} algorithm inspired by previous work~\citep{4766900, chickering2025qlip}. A quadtree is a hierarchical data structure that recursively partitions a surface into four quadrants until a stopping criterion is satisfied. In image processing, quadtrees have been widely used as compact image representations: homogeneous regions are represented by large leaves, while visually complex regions are recursively divided into smaller cells. This property makes quadtrees particularly suitable for GUI screenshots with large background panels, margins as well as small concentrated regions with text and icons. A quadtree can therefore adapt its partition to the information density of the screenshot, using coarse leaves for redundant regions and fine leaves for detail-rich regions.

Based on this observation, \ourmethod first build an adaptive quadtree, and then uses its leaves as adaptive visual-token units, which we describe below. The detailed algorithm and implementation is discussed in \Cref{app-alg}. 

\paragraph{Grid Alignment and Boundary Handling}

Since dimensions of GUI screenshots are not necessarily powers of two, which is a precondition of applying quadtree, \ourmethod therefore first decomposes the image into two parts: a centered region that can be tiled by square chunks of size $C\times C$, and boundary margins that cannot be included in this chunk layout, which are kept untouched. \ourmethod then runs an adaptive quadtree independently inside each chunk. 

\paragraph{Splitting Criteria}

For a quadtree node $n$ of size $w_n\times h_n$, the default split criteria is the area-weighted grayscale variance $s(n)=w_n h_n \cdot \operatorname{Var}(\operatorname{grey}(n))$, and we recursively split until $s(n)\leq\ 1000\cdot \alpha$.

The implementation also supports a gradient-based criterion, where the score function $s$ gives the maximum magnitude of the gradient of node $n$, which can be useful when the edges provide a stronger signal under certain circumstances. We will discuss the choice of different criteria in \Cref{ref-alpha}

\paragraph{Representative Token Selection}
For each final leaf, \ourmethod selects the representative token block. By default, this is the center block $(x' = \lfloor \frac{x_0 + x_1}{2} \rfloor,\quad y' = \lfloor \frac{y_0 + y_1}{2} \rfloor)$ if the leaf spans block coordinates from $x_0$ to $x_1$ and $y_0$ to $y_1$.

\subsection{Conditional Quadtree Building}
\label{method-cond}
GUI trajectories often contain adjacent screenshots that are nearly identical. A page may remain static, a menu may pop up, or a scroll operation may shift content by a few token blocks. If each screenshot is compressed independently, a later frame may discard details that were preserved in an earlier frame, which can weaken the reasoning and grounding ability.

\ourmethod introduces a conditional quadtree mechanism that infers history from the ordered images in the same request. It is a heuristic algorithm that builds quadtree conditioned on the state change of each chunk in one of three modes: static, shifted, or replaced.


\paragraph{Static mode}
We begin by identifying patterns that most pixels remain the same, which we call the Static mode. We define a similarity value between two chunks. Consider the chunk \ourmethod produced previously, we summarize it into a condensed grid signature $S$ of size $(H_c, W_c)$ by computing the greyscale of each block of size $(b,b)$ so that each cell of $S$ stores the rounded mean grayscale of the corresponding $b \times b$ block. $(H_c, W_c)$ denotes the number of blocks within a chunk.

 We define similarity of two adjacent signatures $S_t$ and $S_{t-1}$ as following:
$$
\text{similarity}(S_t,S_{t-1}) = 1 - \frac{1}{255 H_c W_c}\times \sum_{i,j}|S_t(i,j)-S_{t-1}(i,j)| 
$$
A region is static if $\text{similarity} \geq \tau _{static}$.

\paragraph{Shifted mode}
If two chunks are not static, we then consider if a scroll operation happens. We call this the Shifted mode. To capture the scrolling feature, we add a block-level shift $\delta=(\Delta i,\Delta j)$ in the previous definition and change to average over the overlapping area. We search over $|\Delta i|,|\Delta j|$ in certain range and classify the region as shifted if the best valid shift $\delta^\star$ satisfies
\[
\operatorname{sim}(\delta^\star)\ge\tau_{\mathrm{shift}}
\quad\text{and}\quad
\operatorname{sim}(\delta^\star)\ge \operatorname{sim}(S_t, S_{t-1})+\gamma.
\]
\paragraph{Replaced mode}
Regions satisfying neither condition are treated as replaced. A new quadtree is constructed where no priors are used.

We set these parameters heuristically, where $\tau_{static}=0.97, \tau_{shift}=0.94, \gamma=0.03$ and the shift search range is $|\Delta i|,|\Delta j| \leq 4$ blocks. The full algorithm can be seen in \Cref{alg:cond} (\Cref{app-cond}).


\section{Experiment and Results}
We introduce the basic setup of the experiment in \Cref{exp-setup}, and then compare \ourmethod with the baseline in \Cref{exp-main}.  We continue to explore the generalization of \ourmethod on different models with Qwen3-VL architecture. Navigational experiments using AndroidControl and AndroidWorld are discussed in \Cref{exp-navi}.

\subsection{Experiment Setup}
\label{exp-setup}
\textbf{Benchmarks.} To evaluate the performance of \ourmethod, we evaluate it on multiple benchmarks. Thanks to ClawGUI's~\citep{tang2026clawgui} effort on reproducability of benchmarking GUI agent models, we reuse the majority of its experiment setting, which consists of various grounding benchmarks including ScreenSpot-Pro~\citep{li2025screenspot}, ScreenSpot-V2~\citep{wu2024atlas}, OSWorld-g~\citep{xie2024osworld}, UI-Vision~\citep{nayak2025ui}, MMBench-GUI~\citep{wang2025mmbench}, and one offline navigational benchmark, AndroidControl~\citep{li2024effects}. For online navigational benchmark, we choose AndroidWorld~\citep{rawles2024androidworld}. We use grounding benchmarks as the main evaluation because they provide controlled, reproducible measurement, and report navigation benchmarks as complementary interactive evaluations.

\textbf{Models.} We evaluate \ourmethod on the Qwen2-VL~\citep{wang2024qwen2} series and the ShowUI~\citep{lin2025showui} series for a direct comparison with \showui. Additionally, to examine the generalization ability, we evaluate several series of SOTA GUI agent model such as the GUI-Owl-1.5\citep{xu2026mobile}, the MAI-UI~\citep{zhou2025mai} and the Qwen3-VL~\citep{bai2025qwen3} series. These models share the same backbone model as Qwen3-VL but differs with different training strategy.

\textbf{Metrics.} For grounding benchmarks and AndroidControl, we report click accuracy (Acc.). For AndroidWorld, we report success rate (SR). 

\textbf{Baseline.} Our primary baseline is ShowUI's UI-guided token selection, the closest GUI-specific visual-token reduction method with available implementation. We also include a random reduction baseline and FastV in \Cref{ref-layout} to test whether \ourmethod actually benefits from quadtree structure. The model choice is limited by \showui's implementation, as no recent implementation on Qwen3-VL series is provided.  

\subsection{Main Results}
\label{exp-main}
\begin{figure*}[t]
    \centering
    \includegraphics[width=\linewidth]{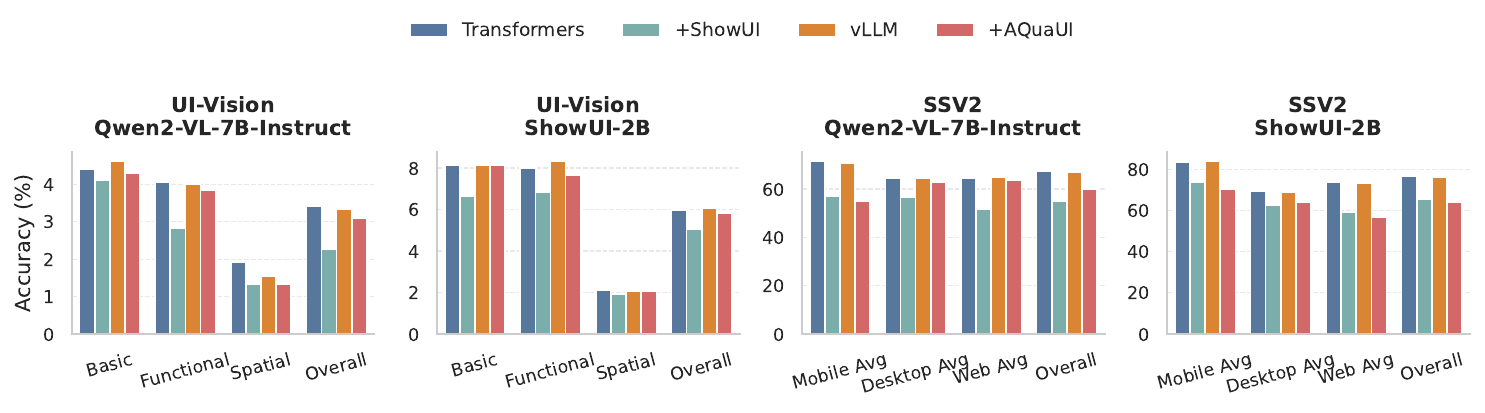}
    \caption{Performance breakdown on UI-Vision and ScreenSpot-V2. More detailed results can be seen in \Cref{app-break}.}
    \label{fig:break}
\end{figure*}

\Cref{tab:main_results} reports the performance and latency differences between \ourmethod and \showui for five grounding benchmarks. \Cref{tab:main_results_other} reports the result of \ourmethod on other SOTA GUI agent models. \Cref{fig:break} displays the performance breakdown on UI-Vision and ScreenSpot-V2. The rest of breakdown performance is reported in detail in \Cref{app-break}. We make the following observations:

\textbf{\ourmethod preserves task accuracy better at comparable compression rates.} At similar compression rates, \ourmethod consistently produces smaller average accuracy degradation than \showui's UI-guided visual token selection. On Qwen2-VL-7B-Instruct, \ourmethod reduces tokens by 26.82\% and merely drops accuracy by 3.29 points on average compared to \showui's 5.95-point drop at 27.55\% compression. A similar trend is observed on ShowUI-2B with a slightly higher compression rate and a smaller average accuracy drop.

\textbf{Token reduction introduces noticable overhead under Qwen2-VL implementations.} Both \showui and \ourmethod introduce additional latency in this setting, despite reducing the number of visual tokens. This suggests that the overhead of token-selection logic on Qwen2-VL can not be neglected, especially for smaller models where the full-token inference cost is already low. Nevertheless, \ourmethod still introduces substantially less overhead on Qwen2-VL-7B-Instruct than \showui, with +0.02s comparing to +0.31s, indicating \ourmethod is more efficient when serving.

\textbf{Spatially structured tasks benefit more from \ourmethod.} \Cref{fig:break} suggests that \ourmethod is particularly effective on tasks requiring preserving structured GUI evidence. On ScreenSpot-V2, \ourmethod introduce ssmaller degradation than ShowUI on Desktop and Web for Qwen2-VL-7B-Instruct, suggesting that \ourmethod better preserves layout-sensitive visual context.



\begin{table*}[t]
\centering
\footnotesize
\setlength{\tabcolsep}{3.5pt}
\renewcommand{\arraystretch}{1.12}
\caption{Accuracy and latency across GUI benchmarks compared with \showui's UI Guided Token Selection. UI\_mask\_ratio is set to default 0.5. Accuracy and compression are reported in percentages; latency is reported in seconds. SS-Pro: ScreenSpot-Pro; SS-v2: ScreenSpot-V2; OSW-G: OSWorld-G; UI-V: UI-Vision; MMB-GUI: MMBench-GUI.}
\begin{tabular}{lccccccccccccc}
\toprule
\multirow{2}{*}{Backend}
& \multicolumn{2}{c}{SS-Pro}
& \multicolumn{2}{c}{SS-V2}
& \multicolumn{2}{c}{OS-G}
& \multicolumn{2}{c}{UI-VISION}
& \multicolumn{2}{c}{MMB-GUI}
& \multirow{2}{*}{Comp.}
& \multicolumn{2}{c}{Avg. $\Delta$} \\
\cmidrule(lr){2-3}
\cmidrule(lr){4-5}
\cmidrule(lr){6-7}
\cmidrule(lr){8-9}
\cmidrule(lr){10-11}
\cmidrule(lr){13-14}
& Acc. & Lat.
& Acc. & Lat.
& Acc. & Lat.
& Acc. & Lat.
& Acc. & Lat.
&
& Acc. & Lat.\\
\midrule

\multicolumn{12}{l}{\textbf{Qwen2-VL-7B-Instruct}} \\
\midrule
Transformers
& 10.31 & 1.10
& 67.30 & 1.03
& 21.14 & 1.25
& 3.41 & 1.17
& 39.57 & 1.32
& 0
& --&--\\

\quad +\showui
& 8.60 & 1.48
& 55.11 & 1.28
& 16.87 & 1.61
& 2.72& 1.42
& 28.69& 1.62
& 27.55
& -5.95& +0.31\\




vLLM
& 10.44 & 1.87
& 67.14 & 1.29
& 19.36 & 1.48
& 3.34 & 1.38
& 39.90 & 1.57
& 0 
& -- &--\\

\quad +\ourmethod
& 9.87& 2.01
& 60.06 & 1.30
& 17.94 & 1.50
& 3.10 & 1.30
& 32.80& 1.59
& 26.82 
& \textbf{-3.29} &\textbf{+0.02}\\


\midrule
\multicolumn{12}{l}{\textbf{ShowUI-2B}} \\
\midrule
Transformers
& 7.53 & 0.74
& 76.65 & 0.61
& 18.29 & 0.64
& 5.97 & 0.65
& 43.49 & 0.66
& 0
& -- & --\\

\quad +\showui
& 5.95 & 0.78
& 65.72 & 0.71
& 16.34& 0.72
& 5.04 & 0.66
& 33.67& 0.76
& 27.78
& -5.04 & \textbf{+0.07}\\



vLLM
& 7.59 & 1.18
& 76.49 & 0.53
& 18.12 & 0.45
& 6.06 & 0.44
& 44.05 & 0.64
& 0 
&-- &--\\

\quad +\ourmethod
& 6.58 & 1.35
& 64.07 & 0.62
& 16.34 & 0.52
& 5.84 & 0.55
& 36.00 & 0.70
& 28.24 
& \textbf{-4.70} & +0.10\\
\bottomrule
\end{tabular}

\label{tab:main_results}
\end{table*}


\textbf{\ourmethod preserves accuracy consistently on Qwen3-VL.} Across the Qwen3-VL series, \ourmethod reduces around 30\% of visual tokens while incurring less than one point of average accuracy degradation. The average drops are only 0.61, 0.48, and 0.85 points for Qwen3-VL-2B, 8B, and 32B, respectively. This suggests that \ourmethod is more robust with Qwen3-VL's positional embedding design.

\textbf{Latency gains emerge as model size increases.} Although token reduction introduces additional overhead, the savings become visible for larger backbones. For Qwen3-VL-8B and Qwen3-VL-32B, \ourmethod reduces average latency by 0.09s and 0.23s. Similar trends appear on other larger models. In contrast, the 2B models show little latency changes. This suggests that inference-time visual-token reduction is especially valuable for larger GUI agents, where the language model computation dominates the reduction overhead.


\label{exp-models}
\begin{table*}[t]
\centering
\small
\setlength{\tabcolsep}{3.5pt}
\renewcommand{\arraystretch}{1.12}
\caption{Accuracy and latency across GUI benchmarks for various GUI agent models. Accuracy and compression are reported in percentages; latency is reported in seconds. SS-Pro: ScreenSpot-Pro; SS-V2: ScreenSpot-v2; OSW-G: OSWorld-G; UI-V: UI-Vision; MMB-GUI: MMBench-GUI.}
\begin{tabular}{lccccccccccccc}
\toprule
\multirow{2}{*}{Backend}
& \multicolumn{2}{c}{SS-Pro}
& \multicolumn{2}{c}{SS-V2}
& \multicolumn{2}{c}{OS-G}
& \multicolumn{2}{c}{UI-VISION}
& \multicolumn{2}{c}{MMB-GUI}
& \multirow{2}{*}{Comp.} 
& \multicolumn{2}{c}{Avg. $\Delta$}\\
\cmidrule(lr){2-3}
\cmidrule(lr){4-5}
\cmidrule(lr){6-7}
\cmidrule(lr){8-9}
\cmidrule(lr){10-11}
\cmidrule(lr){13-14}
& Acc. & Lat.
& Acc. & Lat.
& Acc. & Lat.
& Acc. & Lat.
& Acc. & Lat.
& 
& Acc. & Lat.\\
\midrule

\multicolumn{12}{l}{\textbf{Qwen3-VL-2B-Instruct}} \\
\midrule

vLLM
&  43.96 &  1.17
&  88.84 &  0.43
&  48.49 &  0.31
&  15.24 &  0.38
&  73.34 &  0.50
& 0 
& -- & --\\

\quad +\ourmethod
&  42.88 &  1.17
&  88.84 &  0.46
&  48.49 &  0.31
&  15.06 &  0.41
&  71.56 &  0.52
&  29.52 
& -0.61 & +0.02\\

\midrule

\multicolumn{12}{l}{\textbf{Qwen3-VL-8B-Instruct}} \\
\midrule

vLLM
&  56.23 &  1.95
&  94.26 &  0.74
&  59.68 &  0.52
&  27.78 &  0.63
&  84.31 &  0.86
& 0 
& -- & --\\

\quad +\ourmethod
&  55.28 &  1.67
&  94.26 &  0.71
&  59.33 &  0.52
&  27.72 &  0.58
&  83.28 &  0.77
&  29.72 
& -0.48 & -0.09\\
\midrule

\multicolumn{12}{l}{\textbf{Qwen3-VL-32B-Instruct}} \\
\midrule

vLLM
& 58.57 &  3.24
&  94.89 & 1.29
&  66.25 &  0.97
& 36.17 &  1.35
&  87.15 &  1.51
& 0 
& -- & --\\

\quad +\ourmethod
&  56.17 &  2.44
& 95.05&  1.17
&  64.30 &  1.03
& 35.96 & 1.23
& 87.31 & 1.33
&  29.52
& -0.85 & -0.23\\
\midrule

\multicolumn{12}{l}{\textbf{MAI-UI-2B}} \\
\midrule

vLLM
&  58.89 &  1.25
&  92.45 &  0.50
&  55.24 &  0.37
&  29.99 &  0.41
&  82.64 &  0.57
& 0
& -- & --\\

\quad +\ourmethod
&  57.50 &  1.25
&  91.43 &  0.51
&  55.24 &  0.36
&  29.57 &  0.43
&  81.44 &  0.58
&  29.52
& -0.81 & +0.01\\

\midrule
\multicolumn{12}{l}{\textbf{MAI-UI-8B}} \\
\midrule

vLLM
&  64.52 &  2.15
&  94.81&  1.04
&  64.12&  0.86
&  40.45 &  0.94
&  88.62 &  1.20
& 0 
& -- & --\\

\quad +\ourmethod
&  59.14&  1.88
&  94.18&  0.97
&  63.59 &  0.85
& 39.82 &  0.89
&  87.51 &  1.09
&  29.25
& -1.66 & -0.10 \\
\midrule




\multicolumn{12}{l}{\textbf{GUI-Owl-1.5-8B-Instruct}} \\
\midrule

vLLM
&  68.88 & 2.69
& 93.40 & 0.88
& 67.14 & 0.65
&  37.27 &  0.92
&  82.61 &  1.08
& 0 
& -- & --\\

\quad +\ourmethod
& 67.68 & 2.21
& 93.32 & 0.85
& 64.65 & 0.66
& 36.72 & 0.83
& 81.86 & 0.97
& 29.52 
& -1.01 & -0.10\\

\midrule

\multicolumn{12}{l}{\textbf{GUI-Owl-1.5-32B-Instruct}} \\
\midrule

vLLM
&  70.97 &  3.30
& 94.42&  1.28
&  66.43 &  0.95
& 37.45 & 1.29
& 87.15 & 1.50
&  0
& -- & --\\

\quad +\ourmethod
& 69.89 &  2.77
&  94.03 & 1.16
&  65.90 &  0.89
& 36.72 &  1.11
&  86.53 &  1.30
& 29.52
& -0.67&-0.22\\




\bottomrule
\end{tabular}

\label{tab:main_results_other}
\end{table*}
\subsection{Navigational Results}
\label{exp-navi}

\begin{table*}[t]
\centering
\small
\setlength{\tabcolsep}{3.5pt}
\renewcommand{\arraystretch}{1.12}
\caption{Accuracy for various GUI agent models on AndroidControl.}
\begin{tabular}{lccccccc}
\toprule
\multirow{2}{*}{Backend}
& \multicolumn{2}{c}{High}
& \multicolumn{2}{c}{Low}
& \multirow{2}{*}{Comp.}
& \multicolumn{2}{c}{Avg. $\Delta$}\\
\cmidrule(lr){2-3}
\cmidrule(lr){4-5}
\cmidrule(lr){7-8}
& Acc. & Lat.
& Acc. & Lat.
&
& Acc. & Lat.\\
\midrule

\multicolumn{8}{l}{\textbf{Qwen3-VL-2B-Instruct}} \\
\midrule

vLLM
&  54.80 &  0.67
&  73.83 &  0.66
& 0
& -- & --\\

\quad +\ourmethod
&  54.14 &  0.78
&  73.37 &  0.73
&  29.33 
& -0.56 & +0.09\\

\midrule

\multicolumn{8}{l}{\textbf{Qwen3-VL-8B-Instruct}} \\
\midrule

vLLM
& 56.97  &  1.28
& 66.26  &  1.23
&  0
& -- & --\\

\quad +\ourmethod
&  56.43 &  1.21
&  66.32 &  1.18
&  29.33 
& -0.24 & -0.06\\



\bottomrule
\end{tabular}

\label{tab:androidcontrol_results}
\vspace{-5mm}
\end{table*}

\begin{table*}[t]
\centering
\small
\setlength{\tabcolsep}{3.5pt}
\renewcommand{\arraystretch}{1.12}
\caption{Success rate on AndroidWorld. Random seed is fixed to $30$. Accuracy and compression are reported in percentages.}
\begin{tabular}{lccccc}
\toprule
Backend
& & & &SR.& Comp. \\

\midrule

\multicolumn{3}{l}{\textbf{UI-Voyager}} \\
\midrule

vLLM
&  & & &70.69 & --
 \\

\quad +\ourmethod
&  & & &68.10 &  31.90\\

\quad \quad - Conditional Quadtree
& & &  & 65.52 & 32.06
\\

\midrule

\multicolumn{3}{l}{\textbf{MAI-UI-8B}} \\
\midrule

vLLM
& & &  &57.76 & --
\\

\quad +\ourmethod
& & &  &56.03 & 31.89\\

\quad \quad - Conditional Quadtree
& & &  & 53.44 & 32.06
\\



\bottomrule
\end{tabular}

\label{tab:androidworld_results}
\end{table*}

\Cref{tab:androidcontrol_results,tab:androidworld_results}
evaluate \ourmethod on AndroidControl and AndroidWorld. For AndroidWorld, random seed is fixed to $30$. For MAI-UI-8B's result on AndroidWorld, we report our reproduced dense vLLM baseline under the same evaluation protocol used for \ourmethod.

\textbf{\ourmethod remains stable on AndroidControl.}
On AndroidControl, \ourmethod preserves performance under roughly 29\% visual token compression. For Qwen3-VL-2B-Instruct, the average accuracy drop across high- and low-level instructions is only 0.56 points. This indicates that the quadtree reduction remains effective beyond static grounding benchmarks and transfers to navigational settings.

\textbf{Interactive long-horizon evaluation is more sensitive.}
On AndroidWorld, \ourmethod also maintains most of the dense model performance, but the degradation is larger than on AndroidControl. UI-Voyager drops from 70.69\% to 68.10\%, while MAI-UI-8B drops from 57.76\% to 56.03\%. This is expected because AndroidWorld evaluates multi-step interaction, where small grounding or perception errors may accumulate over the trajectory. Nevertheless, the results suggest that \ourmethod can reduce visual token load while retaining competitive navigation performance.

\textbf{Conditional quadtree building is essential.}
The conditional quadtree mechanism provides a clear benefit on AndroidWorld. Removing it causes a consistent success-rate drop under nearly identical
compression rates. For UI-Voyager, success rate decreases from 68.10\% to 65.52\%; For MAI-UI-8B, success rate decreases from 56.03\% to 53.44\%. This indicates that the gain is not simply due to retaining more tokens. Rather, conditional quadtree refinement helps preserve fine-grained partitions across adjacent screenshots.

\section{Ablation Studies}

\subsection{\ourmethod Preserves UI Layout}
\label{ref-layout}
In this section, we discuss whether \ourmethod's quadtree design is genuinely useful by comparing it with randomly dropping visual tokens~\citep{peng2025can} and FastV~\citep{chen2024image}. For FastV, we retain 65\% of tokens from the fourth transformer layer for a fair comparison, using the implementation from PACT~\citep{dhouib2025pact} and the migration to Qwen3-VL. We point out that FastV prunes inside the transformer and still incurs full prefill cost through early layers, whereas \ourmethod reduces tokens before the language transformer entirely.
\Cref{tab:random_compare} shows that random dropping produces large accuracy drops across all benchmarks, confirming that the adaptive quadtree is the key source of \ourmethod's effectiveness. Especially, random reduction drops ScreenSpot-V2 accuracy from 67.14\% to 57.47\% on Qwen2-VL-7B-Instruct, whereas AQuaUI achieves 60.06\%, recovering 3.77 points over random. FastV performs more steadily than \ourmethod on Qwen2-VL-7B-Instruct as the model can still get full image in early layers. However, it is worth noting that the gap between \ourmethod and FastV is model-dependent. Although FastV performs well on Qwen2-VL-7B-Instruct, it degrades severely on Qwen3-VL-8B-Instruct, falling even below random dropping on ScreenSpot-Pro. 





\begin{table*}[t]
\centering
\small
\setlength{\tabcolsep}{3.5pt}
\renewcommand{\arraystretch}{1.12}
\caption{Accuracy comparison with random token dropping with 30\% visual tokens removed across benchmarks. Accuracy is reported in percentages.}
\begin{tabular}{lccccc}
\toprule
&SS-Pro& SS-V2 & OS-G & UI-Vision & MMB-GUI\\
\midrule
Qwen2-VL-7B-Instruct&&&& & \\
\midrule
vLLM& 10.44& 67.14&19.36&3.34&39.90\\
\quad+Random Reduction &5.44&57.47&15.63&2.65&29.60\\
\quad+FastV  &8.73 & 67.45&14.74&3.30 &36.03\\
\quad+\ourmethod &9.87&60.06&17.94&3.10&32.80\\
\midrule
Qwen3-VL-8B-Instruct&&&& & \\
\midrule
vLLM& 58.57& 94.89&66.25&36.17&87.15\\
\quad+Random Reduction &39.47&86.95&46.18&21.15&74.21\\
\quad+FastV & 25.11&85.85 &40.32 &12.81 &62.91\\
\quad+\ourmethod &55.28&94.26&59.33&27.72&83.28\\
\bottomrule
\end{tabular}

\label{tab:random_compare}
\end{table*}

\subsection{Choice of Compression Parameter and Score Function}
\label{ref-alpha}
In this section, we will discuss how different parameters and score functions affect compression quality. We make the following observations.
\begin{figure*}[tbp]
    \centering
    \includegraphics[width=\linewidth]{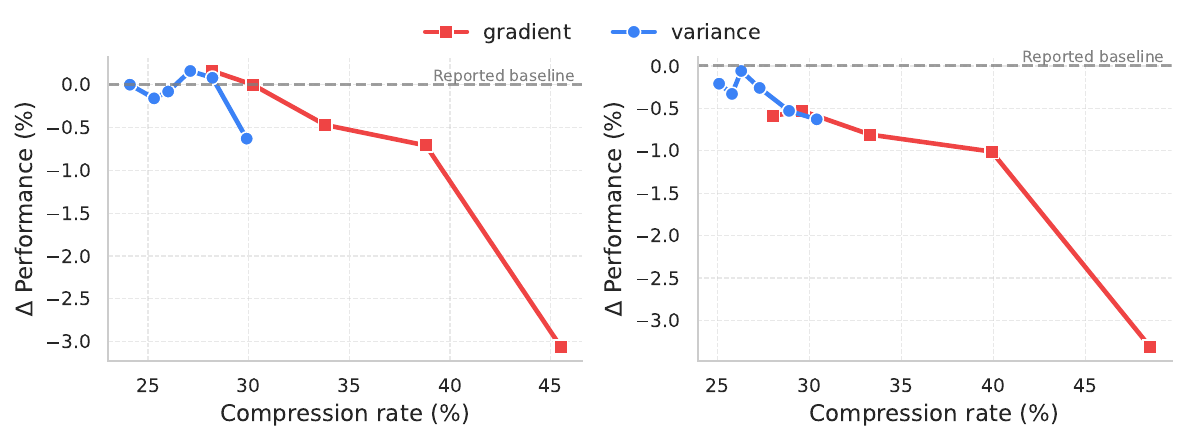}
    \caption{Performance change with respect to different parameter and score function. Left: ScreenSpot-V2. Right: UI-Vision. Each dot represent a choice of $\alpha$. From left-to-right, for variance-based score function, $\alpha=\{1,4,8,16,32,64\}$ and for gradient-based score function, $\alpha=\{10,15,30,60,120\}$.}
    \label{fig:para}
\end{figure*}

\textbf{Variance-based compression is more stable, while gradient-based compression is more aggressive.}
Across ScreenSpot-V2 and UI-Vision, the variance-based strategy yields a relatively narrow compression range and preserves performance consistently. In contrast, the gradient-based strategy reaches substantially higher compression rates, but its performance degrades more noticeably under aggressive compression. 
This suggests a clear stability--compression trade-off between the two strategies. 

\textbf{Appropriate compression can preserve performance with fewer visual tokens.}
The results show that a properly chosen $\alpha$ can reduce visual tokens while maintaining comparable performance. In particular, moderate compression often incurs only a small accuracy change. This indicates that the model tolerates a meaningful amount of visual-token reduction, but excessive compression removes task-relevant visual details.

\section{Conclusion}
We presented \ourmethod, a training-free inference-time visual-token reduction method for GUI agent models. Instead of treating visual-token reduction as a generic token-importance estimation problem, \ourmethod exploits the spatial structure of GUI screenshots through adaptive quadtree partitioning. Homogeneous regions are represented with coarse leaves, while others areas retain finer, with each leaf representing by one visual token. We further introduced a conditional quadtree refinement mechanism that leverages temporal continuity of screenshots, reusing previous partitions only when the current screen region is static or mildly shifted and falling back to reconstruction of a quadtree when the content changes.

Across GUI grounding and navigation benchmarks, \ourmethod consistently reduces visual tokens while preserving accuracy. The results show that spatial redundancy in GUI screenshots can be exploited effectively at inference time, especially for larger GUI agent backbones. More broadly, our findings suggest that efficient GUI perception should not merely ask which tokens are important, but also where the tokens are located.

\clearpage
\bibliographystyle{plainnat}
\bibliography{references}

\appendix


\crefalias{section}{appendix}
\crefalias{subsection}{appendix}
\crefalias{subsubsection}{appendix}

\section{Detailed Algorithm}
\label{app-alg}
\subsection{Grid Alignment and Boundary Handling}
\label{app-chunk}
Given a resized screenshot of size $W \times H$, \ourmethod constructs the adaptive partition on the model's merged-token grid. Let $p$ denote the ViT patch size and $m$ denote the spatial merge size. Each LLM-side visual token corresponds to a block of size $b = p \cdot m$.
For Qwen2-VL and Qwen3-VL, $p=14$ and $m=2$, so $b=28$ pixels.

Since dimensions of GUI screenshots are not necessarily powers of two, directly applying a single quadtree to the whole image won't work. \ourmethod therefore first decomposes the image into two parts: a centered region that can be tiled by square chunks of size $C\times C$, and boundary margins that cannot be included in this chunk layout, which are kept as $b \times b$ token-sized leaves.

On the centered region, \ourmethod tiles square chunks and then runs an adaptive quadtree independently inside each chunk. This design keeps all split boundaries aligned to the merged-token grid and avoids malformed edge regions.

To be more specific, the chunk side length is computed from the maximum depth $d$ supported by the image dimensions:
\[
\begin{aligned}
d_h = \lfloor \log_2(H / b) \rfloor,\quad d_w = \lfloor \log_2(W / b) \rfloor\\
d   = \max(1, \min(d_h, d_w)), \quad C = b \times 2^{(d- 1)}\\
\end{aligned}
\]

\subsection{Splitting Criteria}
For a quadtree node $n$, we will describe the splitting criteria as follows. The default score function $s$ is the area weigth greyscale variance:
$$
s(n) = w_n h_n \cdot \operatorname{Var}(\operatorname{gray}(n)).
$$
A region is recursively split when$s(n) > 1000 \cdot \alpha$.
Area weighting makes the threshold meaningful across different region sizes. Large non-uniform regions are split even when their per-pixel variance is moderate, while large uniform regions can collapse to one token. Increasing alpha therefore increases compression.

The implementation also supports a gradient-based criterion, where the score function $s$ gives the max magnitude of gradient of node $n$, which can be useful when edges provide a stronger signal.

\subsection{Representative Token Selection}

For each final leaf, \ourmethod selects the center merged-token block. If the leaf spans block coordinates from $x_0$ to $x_1$ and $y_0$ to $y_1$, the representative block is$x' = \lfloor \frac{x_0 + x_1}{2} \rfloor,\quad y' = \lfloor \frac{y_0 + y_1}{2} \rfloor$. The representative merged-token coordinate is therefore $(x', y')$

\subsection{Conditional Quadtree Building}
\label{app-cond}
GUI trajectories often contain adjacent screenshots that are nearly identical. A page may remain static, a menu may open, or a scroll operation may shift content by a few token blocks. If each screenshot is compressed independently, a later frame may merge away details that were preserved in an earlier frame, weakening temporal reasoning and action grounding.

\ourmethod introduces a conditional quadtree mechanism that infers history from the ordered images in the same request. We design a heuristic algorithm for this.

\subsubsection{Static, Shifted, and Replaced Modes}

\paragraph{Static mode}

We begin by identify patterns that most pixels remain the same, which we called the Static mode. For this purpose, we define a similarity value between two chunks. Consider the chunks we produced in \Cref{method-quad}, we compress it into a condensed grid signarture $S$ of size $(H_c, W_c)$ by computing the greyscale of each block of size $(b,b)$ so that each cell of $S$ stores the rounded mean grayscale of the corresponding $b \times b$ block with $S \in 
[0,255]^{H_c\times W_C}$. 

For the similarity of two adjacent signatures $S_t$ and $S_{t-1}$, we define similarity as following:
$$
\text{similarity}(S_t,S_{t-1}) = 1 - \frac{1}{255 H_c\cdot W_c}\times \sum_{i,j}|S_t(i,j)-S_{t-1}(i,j)| 
$$

A region is static if $\text{similarity} \geq \tau _{static}$.

\paragraph{Shifted mode}
If two chunks are not static, we then consider if a scroll operation happens. We call this the Shifted mode. To capture the scrolling feature, we add a block-level shift $\delta$ in previous definition. More specifically, given current and previous chunk signatures $S_t,S_{t-1}\in[0,255]^{H_c\times W_c}$, we consider a candidate block-level shift $\delta=(\Delta x,\Delta y)$. We continue to calculate the similarity on the overlapping parts of the two signatures $\Omega(\delta)$. We define $\Omega(\delta)$ as the following:
$$
\Omega(\delta)
=
\{(i,j)\mid
0\le i < H_c,\;
0\le j < W_c,\;
0\le i-\Delta x < H_c,\;
0\le j-\Delta y < W_c
\}.
$$
The overlap ratio is$\rho(\delta)=\frac{|\Omega(\delta)|}{H_cW_c}.$ We only evaluate shifts satisfying $\rho(\delta)\ge\rho_{\min}$.
Therefore, similarity becomes
\[
\text{similarity}(\delta)
=
1-\frac{1}{255|\Omega(\delta)|}
\sum_{(i,j)\in\Omega(\delta)}
\left|S_t(i,j)-S_{t-1}(i-\Delta x,j-\Delta y)\right|.
\]
We search over $|\Delta x|,|\Delta y|\le d_{\max}$ and classify the region as shifted if the best valid shift $\delta^\star$ satisfies
\[
\operatorname{sim}(\delta^\star)\ge\tau_{\mathrm{shift}}
\quad\text{and}\quad
\operatorname{sim}(\delta^\star)\ge \operatorname{sim}(S_t, S_{t-1})+\gamma.
\]
In shifted mode, only coarse previous leaves are translated and reused:
$$
\max(w, h) \ge 2 \times b
$$
Fine leaves are regenerated because small alignment errors around text and icons can be more harmful than independent recompression.
\paragraph{Replaced mode}
Regions satisfying neither condition are treated as replaced. A new quadtree is constructed where no priors are used.

We set these parameters heuristically, where $\tau_{static}=0.97, \tau_{shift}=0.94, \gamma=0.03, \rho_{min}=0.5$ and the shift search range is $|\Delta x|,|\Delta y| \leq 4$ blocks. The full algorithm can be seen in \Cref{alg:cond}

\subsection{Implementation Details}
\label{app-imple}

We implement \ourmethod on both Qwen2-VL and Qwen3-VL with vLLM as the serving backend. 
Although the high-level quadtree construction is shared across models, the two model families require different integration paths because their visual tokenization and position-encoding pipelines differ. 
We describe the model-specific details below.

\subsubsection{Qwen2-VL Details}

Rather than reimplementing Qwen2-VL preprocessing, \ourmethod first runs the vanilla fast processor and then indexes into the resulting dense patch tensor. This preserves Qwen's original resizing, normalization, and patch extraction behavior.

Let $m$ denote the spatial merge size and let $w$ denote the width of the dense merged-token grid. For a representative merged-token coordinate $(x_{\mathrm{rep}}, y_{\mathrm{rep}})$, the corresponding offset in the dense patch tensor is
\[
k = (y_{\mathrm{rep}} \cdot w + x_{\mathrm{rep}}) \cdot m^2 .
\]
The selected patch rows are then
\[
\mathcal{I}_{\mathrm{rep}}
=
\{k, k+1, \ldots, k+m^2-1\}.
\]
For the standard Qwen2-VL setting, $m=2$, so each retained merged token corresponds to four ViT patch rows. This indexing keeps the selected token compatible with Qwen2-VL's original merge-block ordering.

In our default Qwen2-VL implementation, reduction is applied after the vision encoder. The image is first processed by the dense Qwen2-VL ViT, and \ourmethod then selects the visual embeddings corresponding to the representative merged-token coordinates. This conservative path does not reduce ViT computation, but it preserves dense vision encoding quality while reducing the visual sequence passed to the LLM. The reduced visual embeddings are accompanied by their original merged-token coordinates, which are injected into the LLM-side M-RoPE position construction. This avoids treating the sparse selected tokens as a compact dense image grid.

We also implement a pre-ViT Qwen2-VL ablation, where the selected patch rows are passed directly into the vision encoder. This path can reduce both ViT-side and LLM-side computation, but it is more aggressive because the vision encoder receives a sparse visual sequence. We therefore use post-ViT reduction as the default Qwen2-VL setting and treat pre-ViT reduction as a speed-oriented ablation.

\subsubsection{Qwen3-VL Details}

For Qwen3-VL, \ourmethod applies reduction before the vision encoder. After the vanilla fast processor produces the dense patch tensor, we select the patch rows corresponding to the representative quadtree tokens and construct a reduced visual input. Unlike Qwen2-VL post-ViT reduction, this path reduces the number of tokens processed by both the vision encoder and the LLM, allowing larger end-to-end latency gains on Qwen3-VL backbones.

For pre-ViT reduction, the reduced visual tokens still need to be represented by a valid rectangular grid for vLLM's multimodal batching and placeholder expansion. We therefore use aspect-aware padded grid packing: the selected tokens are packed into a near-aspect rectangular grid, and when padding is necessary, the final selected token is duplicated. These duplicate tokens are used only to satisfy shape constraints; their true positions are still tracked through the quadtree metadata.

We patch Qwen3-VL's ViT-side rotary position construction to use true quadtree positions rather than compact packed-grid positions. 
These coordinates follow Qwen's original merge-block-interleaved ordering.

Qwen3-VL also includes an absolute position interpolation path. This path must use the original dense grid extent rather than the maximum retained coordinate. 

Finally, we patch the LLM-side M-RoPE construction. For image tokens, the height and width rotary positions are replaced with the true quadtree coordinates. After injection, the M-RoPE delta is recomputed so that subsequent text tokens remain correctly aligned with the multimodal position sequence. This step is essential: without it, the model would receive fewer visual tokens but would interpret them as occupying compacted positions, which corrupts the geometry needed for GUI grounding.

Qwen3-VL is the more natural target for \ourmethod's pre-ViT reduction, because both vision-side and LLM-side position encodings can be patched to remain consistent with the sparse quadtree layout.

\begin{algorithm}[t]
\caption{Conditional quadtree building for screenshot $s_t$ given $s_{t-1}$}
\label{alg:cond}
\begin{algorithmic}[1]
\Require Current and previous chunks $\{C_t^k\}, \{C_{t-1}^k\}$;
         thresholds $\tau_{\mathrm{static}}, \tau_{\mathrm{shift}}, \gamma, \rho_{\min}, d_{\max}$
\Ensure Refined partition $\mathcal{L}_t$ for $s_t$
\For{each chunk region $k$}
  \State $\mathcal{L}_t^k \gets$ independent quadtree of $C_t^k$
  \State Compute signatures $S_t^k, S_{t-1}^k$
  \If{$\mathrm{sim}(\mathbf{0}) \ge \tau_{\mathrm{static}}$}
    \State $\mathcal{P}^k \gets$ leaves of $C_{t-1}^k$
        \Comment{static}
  \Else
    \State $\delta^{\star} \gets \arg\max_{\delta:\, \rho(\delta) \ge \rho_{\min}} \mathrm{sim}(\delta)$
    \If{$\mathrm{sim}(\delta^{\star}) \ge \tau_{\mathrm{shift}}$
        \textbf{and} $\mathrm{sim}(\delta^{\star}) \ge \mathrm{sim}(\mathbf{0}) + \gamma$}
      \State $\mathcal{P}^k \gets$ coarse leaves of $C_{t-1}^k$ translated by $\delta^{\star}$
        \Comment{shifted}
    \Else
      \State $\mathcal{P}^k \gets \emptyset$
        \Comment{replaced}
    \EndIf
  \EndIf
  \For{each leaf $\ell \in \mathcal{L}_t^k$}
    \State $\mathcal{P}^k(\ell) \gets \{\ell' \in \mathcal{P}^k \mid \ell' \subseteq \ell\}$
    \If{$\mathcal{P}^k(\ell) \neq \emptyset$
        \textbf{and} $\sum_{\ell' \in \mathcal{P}^k(\ell)} \mathrm{area}(\ell') = \mathrm{area}(\ell)$}
      \State Replace $\ell$ with $\mathcal{P}^k(\ell)$ in $\mathcal{L}_t^k$
    \EndIf
  \EndFor
\EndFor
\State \Return $\mathcal{L}_t \gets \bigcup_k \mathcal{L}_t^k$
\end{algorithmic}
\end{algorithm}

\section{Breakdown Statisics on Grounding Benchmarks}
\label{app-break}
\Cref{tab:sspro_breakdown}, \Cref{tab:ssv2_breakdown}, \Cref{tab:osg_breakdown}, \Cref{tab:uivision_breakdown}, \Cref{tab:mmb_breakdown} display the performance breakdown of different benchmarks. We can observe that \ourmethod is particularly effective on tasks requiring preserving structured GUI evidence.
\begin{table}[t]
\centering
\small
\setlength{\tabcolsep}{6pt}
\renewcommand{\arraystretch}{1.12}
\caption{ScreenSpot-Pro breakdown. Accuracy is reported in percentages.}
\begin{tabular}{lccc}
\toprule
Backend & Text & Icon  & Overall \\
\midrule

\multicolumn{4}{l}{\textbf{Qwen2-VL-7B-Instruct}} \\
\midrule
Transformers
& 15.05 & 2.65  & 10.31 \\
\quad + \showui
& 12.28 & 2.65  & 8.60 \\
vLLM
& 15.35 & 2.48  & 10.44 \\
\quad + \ourmethod
& 14.43& 2.48  & 9.87 \\

\midrule
\multicolumn{4}{l}{\textbf{ShowUI-2B}} \\
\midrule
Transformers
& 10.24 & 3.15 & 7.53 \\
\quad + \showui
& 8.09 & 2.48& 5.95 \\
vLLM
& 10.85 & 2.32  & 7.59 \\
\quad + \ourmethod
& 9.31 & 2.15 & 6.58 \\

\bottomrule
\end{tabular}
\label{tab:sspro_breakdown}
\end{table}
\begin{table}[t]
\centering
\small
\setlength{\tabcolsep}{4.5pt}
\renewcommand{\arraystretch}{1.12}
\caption{ScreenSpot-V2 breakdown. Accuracy is reported in percentages.}
\begin{tabular}{lcccccccccc}
\toprule

\multirow{2}{*}{Backend}
& \multicolumn{3}{c}{Mobile}
& \multicolumn{3}{c}{Desktop}
& \multicolumn{3}{c}{Web}
& \multirow{2}{*}{Avg.} \\
\cmidrule(lr){2-4}
\cmidrule(lr){5-7}
\cmidrule(lr){8-10}
& Text & Icon & Avg.
& Text & Icon & Avg.
& Text & Icon & Avg.
& \\
\midrule
\multicolumn{11}{l}{\textbf{Qwen2-VL-7B-Instruct}} \\
\midrule
Transformers
& 75.86 & 65.88 & 71.66
& 74.23 & 51.43 & 64.67
& 69.23 & 58.62 & 64.30
& 67.30 \\

\quad + \showui
& 55.17 & 59.72 & 57.09
& 64.43 & 45.71 & 56.59
& 56.41 & 46.31 & 51.72
& 55.11 \\

vLLM
& 74.14 & 66.35 & 70.86
&73.71 & 51.43 & 64.37
& 69.66 & 59.61 & 64.99
& 67.14 \\

\quad + \ourmethod
& 59.66 & 48.82 & 55.09
& 72.16 & 50.00 & 62.87
& 68.38& 58.13 & 63.62
& 60.06 \\
\midrule
\multicolumn{11}{l}{\textbf{ShowUI-2B}} \\
\midrule
Transformers
& 91.72 & 72.99 & 83.83
& 77.32 & 58.57 & 69.46
& 83.76 & 62.56 & 73.91
& 76.65 \\

\quad + \showui
& 80.00 & 64.93 & 73.65
& 67.53 & 55.71 & 62.57
& 67.09 & 49.75 & 59.04
& 65.72 \\

vLLM
& 91.38 & 74.41 & 84.23
& 76.80 & 57.86 & 68.86
& 82.48 & 63.05 & 73.46
& 76.49 \\

\quad + \ourmethod
& 75.52 & 63.03 & 70.26
& 71.65 & 53.57 & 64.07
& 61.11 & 52.22 & 56.98
& 64.07 \\

\bottomrule
\end{tabular}
\label{tab:ssv2_breakdown}
\end{table}
\begin{table}[t]
\centering
\small
\setlength{\tabcolsep}{6pt}
\renewcommand{\arraystretch}{1.12}
\caption{OSWorld-g breakdown. Accuracy is reported in percentages. Refusal samples are excluded from Overall accuracy. TextMatch: Text Matching; ElemRecog: Element Recognition; LayoutUnd: Layout Understanding; FineManip: Fine-grained Manipulation.}
\begin{tabular}{lccccc}
\toprule
Backend & TextMatch & ElemRecog  &LayoutUnd &FineManip& Overall \\
\midrule

\multicolumn{4}{l}{\textbf{Qwen2-VL-7B-Instruct}} \\
\midrule
Transformers
& 31.67 & 25.16  & 22.59 & 16.67 & 21.14 \\
\quad + \showui
& 25.83 & 19.61  & 17.99 &12.12 &16.87 \\
vLLM
& 28.75 & 22.55  & 21.34 & 16.67 & 19.36 \\
\quad + \ourmethod
& 27.92& 21.57  & 18.41& 13.64&17.94\\

\midrule
\multicolumn{4}{l}{\textbf{ShowUI-2B}} \\
\midrule
Transformers
& 31.67 & 23.53 & 22.18 &14.39 & 18.29\\
\quad + \showui
& 26.25 & 22.22& 19.25 &11.36 &16.34 \\
vLLM
& 31.67 & 23.86  & 21.76 &12.88 & 18.12 \\
\quad + \ourmethod
&27.50 & 21.24 &19.67 & 12.12 & 16.34\\

\bottomrule
\end{tabular}
\label{tab:osg_breakdown}
\end{table}
\begin{table}[t]
\centering
\small
\setlength{\tabcolsep}{6pt}
\renewcommand{\arraystretch}{1.12}
\caption{UI-Vision breakdown. Accuracy is reported in percentages.}
\begin{tabular}{lcccc}
\toprule
Backend & Basic & Functional  & Spatial & Overall \\
\midrule

\multicolumn{4}{l}{\textbf{Qwen2-VL-7B-Instruct}} \\
\midrule
Transformers
& 4.40 & 4.06  & 1.91 & 3.41  \\
\quad + \showui
& 4.12 & 2.82  & 1.34 &2.27 \\
vLLM
& 4.63 & 4.01  & 1.55 & 3.34  \\
\quad + \ourmethod
& 4.29& 3.84  & 1.34& 3.10\\

\midrule
\multicolumn{4}{l}{\textbf{ShowUI-2B}} \\
\midrule
Transformers
& 8.13 & 8.01 & 2.12 &5.97 \\
\quad + \showui
& 6.66 & 6.83& 1.91 &5.04  \\
vLLM
& 8.13 & 8.35  & 2.07 &6.06 \\
\quad + \ourmethod
& 8.13 & 7.67 & 2.07 & 5.84 \\

\bottomrule
\label{tab:uivision_breakdown}
\end{tabular}

\vspace{10pt}

\caption{MMBench-GUI (L2 Grounding) breakdown. Accuracy is reported in percentages.}
\begin{tabular}{lcccccccc}
\toprule
Backend & Windows & MacOS  & Linux & iOS & Android & Web & Overall \\
\midrule

\multicolumn{4}{l}{\textbf{Qwen2-VL-7B-Instruct}} \\
\midrule
Transformers
& 25.41  & 26.63  & 21.71 & 59.01 & 52.18 & 42.88 & 39.57  \\

\quad + \showui
& 19.52  & 14.91 & 18.09 & 41.15 &39.24 &33.66 & 28.69  \\

vLLM
& 26.70  & 27.79  & 20.93 & 58.54 & 51.90 & 43.69 & 39.90  \\

\quad + \ourmethod
& 23.57 & 26.63  & 19.64 & 43.63 & 34.88 &42.39 &32.80  \\

\midrule
\multicolumn{4}{l}{\textbf{ShowUI-2B}} \\
\midrule
Transformers
& 28.36 & 30.54 & 19.12 &63.82 &56.96 & 49.84 & 43.49\\
\quad + \showui
& 21.92 & 18.38& 17.05 &50.62 & 45.85 & 39.81 & 33.67  \\
vLLM
& 29.28 & 31.26  & 19.38 &64.29 & 56.68 &51.13 & 44.05 \\
\quad + \ourmethod
&24.86 &27.35 &10.08 & 51.55 &47.26 & 42.56 & 36.00 \\

\bottomrule
\label{tab:mmb_breakdown}
\end{tabular}
\end{table}

\section*{Limitations}
\label{app-limit}

\paragraph{Model- and serving-stack-specific implementation.}
\ourmethod is training-free, but it is not implementation-free. Correctly preserving visual-token positions requires modifying model-specific position-encoding paths and transporting quadtree metadata through the serving stack. Our current implementation focuses on Qwen2-VL and Qwen3-VL in vLLM. Adapting the same idea to other multimodal models may require additional engineering, especially when the model uses different visual token layouts, position encodings, placeholder expansion rules, or multimodal batching interfaces.

\paragraph{Limited gains when reduction overhead dominates.}
Although \ourmethod reduces the number of visual tokens, it also introduces preprocessing overhead from quadtree construction, metadata transport, and position injection. As a result, latency gains are not guaranteed for small models or settings where the vision encoder and LLM-side computation are already inexpensive. Our results show that the speedup becomes more visible on larger backbones, where the saved visual-token computation can outweigh the extra reduction overhead. More importantly, there are operations in our algorithm that can vectorized for speedup, but we don't implement these improvements for now.

\paragraph{Qwen2 and Qwen3 reduction paths differ.}
The method is more naturally supported on Qwen3-VL, where pre-ViT reduction can be combined with patched ViT-side and LLM-side position encodings. For Qwen2-VL, our conservative default applies reduction after the vision encoder, which preserves dense visual encoding quality but does not reduce ViT computation. Therefore, the latency benefit on Qwen2-VL is more limited. This difference suggests that the effectiveness of serving-time token reduction depends not only on the compression algorithm, but also on where the model architecture allows sparse visual tokens to be introduced safely.

\section*{Broader Impacts}
\label{app-impact}

This work aims to improve the efficiency of GUI agents by reducing redundant visual tokens in high-resolution screenshots. More efficient GUI-agent can lower inference cost and reduce GPU memory usage. This could make screen-grounded agents more practical on edge devices, which may support broader access to mobile-device control and cross-platform user-interface interaction.

\end{document}